%% file: main.tex
\documentclass[11pt]{article}
\input{preamble.tex}

\title{Pointer-Augmented Autoregressive Generation of Patent\\ Claims with Joint Topology and Content Decoding}

\author{
Yongmin Yoo, Zhangkai Wu, Longbing Cao \\
Frontier AI Research Centre, Macquarie University \\
School of Computing, FSE, Macquarie University\\
\texttt{yooyongmin91@gmail.com} \quad  \\
\texttt{\{zhangkai.wu, longbing.cao\}@mq.edu.au}
}

\begin{document}
\maketitle
\input{sec0_abstract.tex}
\input{sec1_intro.tex}
\input{sec2_related_work.tex}
\input{sec3_methodology.tex}

\input{sec5_experiments.tex}
\input{sec6_result}
\input{sec7_conclusion.tex}
\input{sec8_limitation_ethics.tex}

\bibliography{anthology,custom}
\bibliographystyle{acl_natbib}

\appendix
\input{sec8_appendix.tex}

\end{document}

%% file: preamble.tex
\pdfoutput=1

\usepackage[T1]{fontenc}
\usepackage[utf8]{inputenc}
\usepackage{times}
\usepackage{latexsym}
\usepackage{inconsolata}
\usepackage[final]{ACL2023} 

\usepackage{microtype}

\usepackage{amsmath}
\usepackage{amssymb}
\usepackage{amsthm}

\usepackage{algorithm}
\usepackage{algpseudocode}

\usepackage{booktabs}
\usepackage{multirow}
\usepackage{graphicx} 
\usepackage{graphicx}
\usepackage{tikz}
\usepackage{tabularx} 

\usepackage{xcolor}

\usepackage{hyperref}
\usepackage[capitalize,nameinlink]{cleveref}

\usepackage{natbib}

\usepackage{xcolor}
\usepackage{array} 
\usepackage{tabularx} 
\usepackage{multirow}  
\usepackage{float}  
\usepackage{fancyvrb}
\usepackage{tabularx}  
\usepackage{threeparttable}
\usepackage{makecell}
\usepackage{booktabs}
\usepackage{adjustbox}
\usepackage{tcolorbox}

\usepackage{amsmath}

\usepackage{listings}
\usepackage{xcolor}
\input{math}

%% file: math.tex
\newcommand{\mathbold}[1]{\ensuremath{\boldsymbol{\mathbf{#1}}}}

\newcommand{\nestedmathbold}[1]{{\mathbold{#1}}}

\newcommand{\mbe}{\nestedmathbold{e}}

\newcommand{\mbv}{\nestedmathbold{v}}

\newcommand{\mby}{\nestedmathbold{y}}

\newcommand{\cD}{\mathcal{D}}

\newcommand{\cF}{\mathcal{F}}

\newcommand{\cT}{\mathcal{T}}
\newcommand{\cV}{\mathcal{V}}
\newcommand{\cE}{\mathcal{E}}

%% file: sec0_abstract.tex
\begin{abstract}
Autoregressive decoders emit flat token sequences and cannot enforce hierarchical constraints across output segments, a limitation that becomes acute in patent claim generation, where a claim set forms a dependency forest whose scope must narrow monotonically with depth. Topology and content are mutually dependent: a dependent claim's wording must reflect its parent's scope, yet the parent must be chosen before that wording exists, so neither post-hoc parsing nor grammar-constrained decoding suffices. We propose SPG (Structure-aware Patent Generation), which predicts topology \emph{inside} the autoregressive pass. A pointer head selects each dependent claim's parent, and its gradients, together with a depth-adaptive scope regularizer, reshape the shared decoder's representations during training. A second stage then applies a violation-weighted preference objective over self-generated deficient candidates, supplying the negative signal that granted-patent corpora lack. On HUPD-DCG, SPG on Llama-3-8B-Instruct recovers 79.0\% of gold parent links, a quantity its training reward never supervises, and raises antecedent consistency from 0.292 to 0.478 over a supervised baseline of equal scale, with expert evaluation corroborating these gains.
\end{abstract}

%% file: sec1_intro.tex
\section{Introduction}

Autoregressive language models generate text as a flat sequence of tokens, without explicit mechanisms for reasoning about hierarchical relationships among output segments~\citep{li2024pre}. Yet many generation tasks produce outputs that are inherently tree-structured, from abstract syntax trees in code generation~\citep{yin2017syntactic,rabinovich2017abstract} to logical-form trees in semantic parsing~\citep{dong2016language}. Such tasks require the decoder to jointly predict the tree topology and ensure that the content of each node is semantically consistent with its ancestors. The challenge is most acute in domains where the hierarchy carries formal constraints that are legally or programmatically enforceable, since violations there cannot be dismissed as mere disfluencies.

Patent claim generation is a demanding instance of this challenge. A claim set forms a \emph{Claim Dependency Forest} in which each independent claim roots a tree and each dependent claim cites an antecedent through hierarchical reference~\citep{faber2015mechanics}. What distinguishes this setting from tree-structured generation in general is that the hierarchy is statutory: the scope of protection must narrow monotonically with depth, and failure to do so is a ground for rejection~\citep{mpep2173}. The same formality makes the domain an ideal testbed, as the tree structure is explicitly annotated, the violation criteria are defined by statute, and compliance can be assessed automatically~\citep{yoo-etal-2025-patentscore}.

Existing approaches generate claims individually or concatenate them into a flat sequence. Two architectural limitations follow. First, standard autoregressive decoding lacks the capacity to model non-linear parent\,/\,child dependencies among claims~\citep{liu2022autoregressive}, and frequently produces broken dependency chains~\citep{jiang-etal-2025-hupd-dcg,zuo2024patenteval}. Second, no existing method supplies an inductive bias for scope narrowing across depths~\citep{faber2015mechanics}, so models generate dependent claims that inadvertently broaden or contradict their parent's scope. A third limitation is independent of architecture: aligning a model with these constraints requires contrastive examples of deficient outputs, yet public patent corpora consist almost exclusively of granted applications~\citep{suzgun2023hupd,casola2022summarization}, leaving no negative signal to learn from.

The first two limitations share a common root: \textit{structural decisions are made outside the generation process}. Prior structured decoders~\citep{alvarez2017tree} predict topology in isolation from content, while grammar-constrained methods~\citep{geng2023grammar} enforce surface well-formedness without semantic consistency across depths. If topology is instead predicted \emph{within} the autoregressive pass, the structural decision and the token stream share a single set of representations, and gradients from the former reshape the latter during training. Tree shape and node content then co-adapt, which neither post-hoc parsing nor constrained decoding can achieve. The third limitation calls for a different remedy, namely manufacturing the negative signal that the corpus withholds.

\textbf{SPG} (\textbf{S}tructure-aware \textbf{P}atent \textbf{G}eneration) follows from a single commitment: the representation that emits a claim's tokens must also be the one that decides where the claim attaches. A pointer head at each dependent claim's opening delimiter therefore reads the very hidden states used for token prediction, so its supervision flows back into them and leaves the decoder itself topology-aware. Attachment alone, however, fixes only \emph{which} claim a child cites, not what that citation obliges it to say, so we constrain the parent-child relation directly through a margin whose radius decays geometrically with depth, a constraint that is inert until the pointer objective makes such pairs exist. Both objectives, being teacher-forced on granted patents, never show the model what a defective forest costs; Legal Preference Optimization supplies that signal from the policy's own deficient samples, scaled by severity because uniform weighting penalises a missing semicolon and a missing subtree alike, and collapses. \looseness=-1

We evaluate SPG on HUPD-DCG~\citep{jiang-etal-2025-hupd-dcg}. Our
contributions are as follows.
\begin{itemize}
    \item \textbf{Joint decoding.} A pointer head predicts each dependent claim's parent inside the autoregressive pass, so topology and content are learned in one forward pass without a separate parsing stage.

    \item \textbf{Scope regularization.} A depth-adaptive margin with geometrically decaying radii tightens parent/child coupling with depth, acting on the very pairs the pointer objective brings into existence.

    \item \textbf{Severity-weighted alignment.} Scaling each preference pair by the deficiency of the rejected sample averts the mode collapse that uniform weighting induces, beyond what a magnitude-matched control explains.

    \item \textbf{Empirical dissociation.} Across 6B to 141B parameters, fluency scales while structural compliance does not, and domain pretraining closes no part of the gap, indicating that hierarchical competence requires an explicit inductive bias rather than capacity.
\end{itemize}

%% file: sec2_related_work.tex
\section{Related Work}

\subsection{Tree-Structured Autoregressive Generation}
\label{sec:rw_structure}
Tree-structured outputs have been generated autoregressively by doubly-recurrent decoders~\citep{alvarez2017tree} and coarse-to-fine strategies~\citep{dong2018coarse} for code generation and semantic parsing, by tree-aware positional encodings that inject hierarchical bias into Transformers~\citep{shiv2019novel}, and by stack-pointer architectures that resolve non-local dependencies in parsing~\citep{ma2018stackptr}. On the decoding side, grammar-constrained methods instead guarantee well-formedness by restricting the output vocabulary at each step~\citep{geng2023grammar}. These lines share a separation of concerns: tree decoders and pointer parsers produce structurally valid outputs without enforcing semantic consistency across depths, while grammar constraints are agnostic to inter-node coherence. SPG removes that separation by routing the pointer's gradients through the same decoder that emits the tokens, and by adding a depth-adaptive margin that couples a child's representation to its parent's, so topology and cross-depth semantics are learned in one pass.

\subsection{Patent Document Generation}
\label{sec:rw_patent}
Patent generation began by treating each claim as an isolated sequence, fine-tuning GPT-2 on USPTO claims~\citep{lee2023patentgptj} and extending this to multiple sections through a prompt-based multi-task model~\citep{christofidellis2022pgt}; \citet{casola2022summarization} survey the broader space of patent summarization, simplification, and generation. Recent work conditions on richer input, showing that description-based claim generation substantially outperforms abstract-based methods~\citep{jiang-etal-2025-hupd-dcg}, a finding that carries over to European data~\citep{jiang2025enriching}, while error analyses identify broken dependency chains and missing antecedent bases among the most frequent failure modes~\citep{zuo2024patenteval}. Adjacent efforts combine knowledge-graph pre-training with RLHF for inventive concept articulation~\citep{ren2025patentgpt} and target document-level drafting through outline-guided and graph-based generation~\citep{knappich2025pap2pat,yoo2026flowplan}, largely over HUPD~\citep{suzgun2023hupd}. All of these emit claims individually or as one flat concatenation; none predicts the topology of a claim set or enforces the statutory requirement that a dependent claim narrow its antecedent's scope, which is precisely the gap we address.

\subsection{Preference Optimization for Language Model Alignment}
\label{sec:rw_preference}
Preference-based alignment has moved from reward-model RLHF~\citep{ouyang2022training} to objectives that bypass explicit reward modeling: DPO reparameterizes the reward as an implicit log-likelihood ratio~\citep{rafailov2023direct}, IPO replaces the Bradley-Terry assumption with a general preference likelihood~\citep{azar2024general}, and KTO operates on unpaired binary feedback~\citep{ethayarajh2024kto}, with applications concentrated on helpfulness and safety~\citep{xu2024dpo}. Adapting these objectives to domain-specific formal constraints remains largely unexplored; recent DPO work on medical vision-language models~\citep{kim2026dpomedical}, for instance, adopts the standard objective unchanged. LPO departs from this on two points: it scales the contrastive loss by the degree of statutory violation, so that severity rather than mere preference determines gradient magnitude, and it synthesises rejected candidates from the current policy instead of drawing on a fixed offline preference set.

%% file: sec3_methodology.tex
\section{Methodology}
\label{sec:method}

\begin{figure*}[t]
\centering
\includegraphics[width=0.99\textwidth,keepaspectratio]{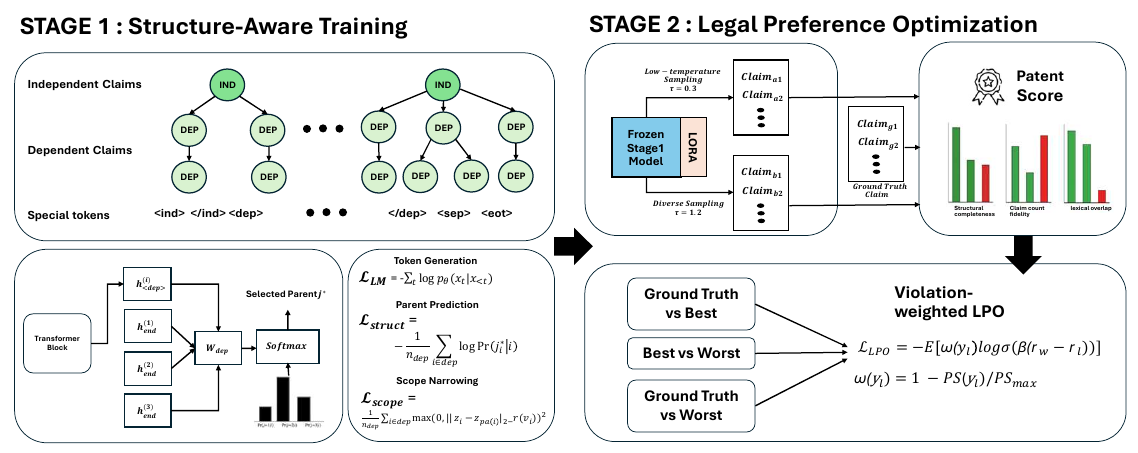}
\caption{Overview of SPG. \textbf{Stage~1}: \emph{Tree-Structured Decoding} (\S\ref{sec:tree_decoding}) jointly generates tokens and predicts parent dependencies via a bilinear pointer ($W_{\mathrm{dep}}$), supervised by structure consistency loss $\mathcal{L}_{\texttt{struct}}$ and depth-adaptive scope regularization $\mathcal{L}_{\texttt{scope}}$. \textbf{Stage~2}: \emph{Legal Preference Optimization} (\S\ref{sec:lpo}) generates candidates at two temperatures, scores them with $\mathrm{PS}(\cdot)$, and applies a violation-weighted objective $\mathcal{L}_{\texttt{LPO}}$ that scales gradients by deficiency severity.}
\label{fig:framework_overview}
\vspace{-4mm}
\end{figure*}
 
Given a patent description~$\cD$, SPG generates a claim set~$\mathcal{C}$ organized as a Claim Dependency Forest~$\cF$. We first formalize the output structure, then describe the three components of the framework: pointer-augmented decoding (\S\ref{sec:tree_decoding}), auxiliary structural objectives (\S\ref{sec:multitask}), and legal preference optimization (\S\ref{sec:lpo}). Training proceeds in two stages: Stage~I fine-tunes the base model with the structure-aware objectives in \S\ref{sec:tree_decoding} and \S\ref{sec:multitask}; Stage~II applies preference optimization over the Stage~I policy (\S\ref{sec:lpo}).\looseness=-1

% ============================================================
\subsection{Tree-Structured Decoding}
\label{sec:tree_decoding}
% ============================================================

\paragraph{Claim Dependency Forest.}
A patent claim set comprises independent claims, each defining a distinct inventive concept, and dependent claims, each referencing exactly one antecedent claim. We formalize this as a forest $\cF=\{\cT_1,\dots,\cT_k,\dots,\cT_K\}$ where each tree $T_k = (V_k, E_k)$ is rooted at an independent claim $\mbv_{\mathrm{root}}^k$, and an edge $(\mbv_i, \mbv_j) \in E_k$ indicates that claim~$j$ depends on claim~$i$. Refer to~\Cref{tab:notation} for detailed symbols and meanings.

\paragraph{Linearization.}
We serialize $\mathcal{F}$ in depth-first order with children sorted by claim number. Structural delimiters \texttt{<ind>}/\texttt{</ind>} and \texttt{<dep>}/\texttt{</dep>} mark claim boundaries; \texttt{<sep>} separates trees and \texttt{<eot>} terminates the forest. For each tree $\cT_k$, we denote its root as $\mbv_{\mathrm{root}}^k$ and let $\operatorname{DFS}_k = \operatorname{DFS}(\cT_k \setminus \{\mbv_{\mathrm{root}}^k\})$ denote the depth-first traversal order of all non-root nodes in $\cT_k$. In words, each tree is written as its independent (root) claim followed by its dependent claims in depth-first order, and the trees are concatenated with a separator between them and an end marker after the last, giving a single token sequence $\mathcal{S}$:

\begin{align}
  \mathcal{S}_k 
  &= \texttt{<ind>}\oplus \mbv_{\mathrm{root}}^k
       \oplus\texttt{</ind>} \nonumber\\
  &\quad \oplus\!\bigoplus_{j\in\operatorname{DFS}_k}\!
       \bigl(\texttt{<dep>}\oplus \mbv_j
       \oplus\texttt{</dep>}\bigr),
  \label{eq:serial}\\[2pt]
  \delta_k
  &=
  \begin{cases}
    \texttt{<sep>}, & k\in\{1,\ldots,K-1\},\\
    \texttt{<eot>}, & k=K.
  \end{cases}
  \nonumber
\end{align}
where $\mathcal{S}:= \bigoplus_{k=1}^{K}\bigl(\mathcal{S}_k \oplus \delta_k\bigr)$.

\paragraph{Pointer-based parent selection.}

The key design choice is to couple the dependency decision with generation through a \emph{shared backbone}: rather than re-injecting a parent vector at inference, the pointer head reads the same hidden states used for token prediction, so its training gradients reshape those representations to be topology-aware (\S\ref{sec:multitask}). We augment the autoregressive decoder with an explicit dependency selection mechanism. Let $f(\cdot\,;\theta)$ denote the causal language model that maps the serialized forest to a sequence of hidden states $\{\mathbf{h}_t\}_{t=1}^{|\mathcal{S}|}=f(\mathcal{S};\theta)$. At the opening delimiter \texttt{<dep>} of each dependent claim~$\mbv_i$, the model selects its parent from the candidate set $\mathcal{A}(i)=\{\mbv_j\in V_k : j \text{ precedes } i \text{ in } \mathcal{S}_k\}$. The dependency score is computed with a learnable bilinear projection $W_{\mathrm{dep}}\!\in\!\mathbb{R}^{d\times d}$, yielding the selection distribution:
\begin{equation}
  \Pr(j \mid i) =
    \frac{\mbe(i,j)}
         {\sum_{j'\in\mathcal{A}(i)} \mbe(i,j')},
  \label{eq:pointer}
\end{equation}
where $\mbe(i,j)=\exp\!\bigl(\mathbf{h}_{\texttt{<dep>}}^{(i)\top} W_{\mathrm{dep}}\,\mathbf{h}_{\texttt{end}}^{(j)}\bigr)$. Here $\mathbf{h}_{\texttt{<dep>}}^{(i)}$ is the hidden state at the dependent claim's opening delimiter and $\mathbf{h}_{\texttt{end}}^{(j)}$ is the hidden state at each candidate's closing delimiter. The parent-selection step is thus jointly parameterized by $(\theta, W_{\mathrm{dep}})$: the language model parameters~$\theta$ produce contextualized representations, while $W_{\mathrm{dep}}$ learns a task-specific compatibility function over those representations. To stabilize early training, we initialize $W_{\mathrm{dep}}$ as an identity matrix scaled by $1/\sqrt{d}$, which reduces the initial structure loss magnitude and prevents the pointer objective from dominating the language modeling gradient. The pointer is supervised with cross-entropy and decoded greedily at inference.

%============================================================
\subsection{Auxiliary Objectives}
\label{sec:multitask}
% ============================================================

Two auxiliary losses complement the language modeling
objective in Stage~I.

\paragraph{Structure consistency loss.}
The pointer distribution in~\Cref{eq:pointer} is supervised with cross-entropy against the ground-truth parent~$j_i^*$:
\begin{equation}
  \mathcal{L}_{\texttt{struct}} =
  -\frac{1}{n_{\texttt{dep}}}
  \sum_{i \in \texttt{dep}}
  \log\,\Pr(j_i^*\mid i).
\end{equation}

\paragraph{Hierarchical scope regularization.}
We encourage each dependent claim's mean-pooled representation $\mathbf{z}_i=\frac{1}{|\mbv_i|} \sum_t\mathbf{h}_t^{(i)}$ to remain close to that of its parent:\looseness=-1
\begin{multline}
  \mathcal{L}_{\texttt{scope}} =
  \frac{1}{n_{\texttt{dep}}}
  \sum_{i\in\texttt{dep}}
  \max\bigl(0,\;
  \|\mathbf{z}_i-\mathbf{z}_{\mathrm{pa}(i)}\|_2 \\
  - \mathrm{r}(\mbv_i)
  \bigr)^{2},
\label{eq:scope}
\end{multline}
where $d(\mbv_i)\in\mathbb{N}$ denotes the depth of claim~$\mbv_i$ in its dependency tree (the root has depth~$1$), $\rho>0$ is the base radius, and $\lambda\in(0,1)$ is a geometric decay factor, so that the allowed radius $\mathrm{r}(\mbv_i)=\rho\cdot\lambda^{d(\mbv_i)-1}$ shrinks with increasing depth.

% ============================================================
\subsection{Legal Preference Optimization}
\label{sec:lpo}
% ============================================================

\paragraph{Preference construction.}
We define a deterministic scoring function $\operatorname{PS}(\mby)\in[0,4.5]$ that scores a generated claim set against the reference through five rule-based sub-metrics: \emph{Claim Count Match} (max 1.5), \emph{Structural Pattern} (max 1.0), \emph{Antecedent Consistency} (max 1.0), \emph{Length Ratio} (max 0.5), and \emph{Content Overlap} (max 0.5), together covering claim-count fidelity, structural well-formedness, referential validity, and lexical overlap; each is deterministic, format-neutral, and cost-free, with the full specification and calibration in \Cref{app:patentscore}. The structure-aware fine-tuned policy (\S\ref{sec:tree_decoding}) generates candidates for $B{=}2{,}000$ training descriptions at two sampling temperatures ($\tau{=}0.3$ and $\tau{=}1.2$). Each candidate is scored by $\mathrm{PS}(\cdot)$, and up to three pairs are constructed per description (not all descriptions satisfy the margin conditions), yielding 5{,}417 preference triples of three types: \emph{gt-vs-worst}, pairing the ground-truth with the lowest-scored generation; \emph{best-vs-worst}, pairing the highest- and lowest-scored generations when their gap exceeds a margin~$\kappa$; and \emph{gt-vs-best}, pairing the ground-truth with the best generation when a score gap exceeds a threshold~$\delta$. All triples are stored in a static preference set~$\mathcal{P}$.

\paragraph{Violation-weighted objective.}
Our objective is a standard DPO loss with one change: each preference pair is scaled by a weight $\omega(\mby_l)$ that grows as the rejected sample $\mby_l$ becomes more deficient, so that grossly deficient outputs contribute larger gradients than near-correct ones. Concretely, we set
$\omega(\mby_l)=1-\mathrm{PS}(\mby_l)/\mathrm{PS}_{\max}$:
\begin{equation}
\begin{aligned}
  \mathcal{L}_{\texttt{LPO}} &=
  -\mathbb{E}_{(\cD,\mby_w,\mby_l)}\bigl[
    \omega(\mby_l)\\
    &\log\sigma\!\bigl(
      \beta\,\mathrm{r}_\theta(\mby_w,\cD)
      - \beta\,\mathrm{r}_\theta(\mby_l,\cD)
    \bigr)
  \bigr],
\end{aligned}
  \label{eq:lpo}
\end{equation}
where $\mathrm{r}_\theta(\mby,\cD)= \log\frac{\pi_\theta(\mby\mid \cD)} {\pi_{\texttt{ref}}(\mby\mid \cD)}$ is the implicit reward and $\beta$ controls KL regularization strength.

\paragraph{Overall objective.}
Training proceeds in two stages:
\begin{align}
  \mathcal{L}^{\texttt{I}}  &= \mathcal{L}_{\texttt{LM}}
    + \gamma\,\mathcal{L}_{\texttt{struct}}
    + \eta\,\mathcal{L}_{\texttt{scope}};
  \label{eq:stage1}\\[2pt]
  \mathcal{L}^{\texttt{II}} &= \mathcal{L}_{\texttt{LPO}}.
  \label{eq:stage2}
\end{align}
Here $\mathcal{L}_{\texttt{LM}} = -\sum_{t}\log p_\theta(x_t \mid x_{<t})$ is the standard next-token cross-entropy loss over the serialised forest sequence~$\mathcal{S}$. In Stage~I, the model is trained with $\mathcal{L}^{\texttt{I}}$ to establish coherent claim syntax and structural consistency. In Stage~II, a LoRA adapter is attached to the frozen Stage~I checkpoint and fine-tuned with $\mathcal{L}_{\texttt{LPO}}$ alone; the structural losses are disabled ($\gamma=\eta=0$) and embeddings are frozen. The reference policy $\pi_{\texttt{ref}}$ is fixed to the Stage~I checkpoint throughout Stage~II.

%% file: sec5_experiments.tex
% ============================================================
% Main table
% ============================================================
\begin{table*}[h]
\centering
\small
\setlength{\tabcolsep}{15pt} 
\begin{tabular}{@{}llcccccc@{}}
\toprule
\textbf{Model} & \textbf{Params}
  & \textbf{BLEU} & \textbf{R-1}
  & \textbf{BS} & \textbf{SP} & \textbf{AC} & \textbf{CC} \\
\midrule
\multicolumn{8}{@{}l}{\emph{Domain-specific LLMs}} \\
PatentGPT-J-6B       & 6B   & 12.86 & 30.68 & 80.24 & 0.162 & 0.176 & 0.060 \\
SaulLM-7B            & 7B   & 12.68 & 36.63 & 83.13 & 0.139 & 0.169 & 0.069 \\
\midrule
\multicolumn{8}{@{}l}{\emph{General-purpose LLMs (Medium Size)}} \\
Mistral-7B           & 7B   & 29.70 & 49.17 & 85.33 & 0.165 & 0.183 & 0.008 \\
Llama-3-8B           & 8B   & 35.42 & 58.25 & 88.54 & 0.151 & 0.163 & 0.084 \\
Qwen-3.5-9B          & 9B   & 19.13 & 53.19 & 84.01 & 0.754 & 0.350 & 0.534 \\
\midrule
\multicolumn{8}{@{}l}{\emph{General-purpose LLMs (Large Size)}} \\
Llama-3-70B          & 70B  & 36.40 & 59.89 & 87.44 & 0.184 & 0.214 & 0.072 \\
Mixtral-8$\times$7B  & 47B  & 37.03 & 60.18 & 88.51 & 0.171 & 0.178 & 0.091 \\
Mixtral-8$\times$22B & 141B & 33.96 & 60.57 & 88.97 & 0.221 & 0.192 & 0.068 \\
\midrule
\multicolumn{8}{@{}l}{\emph{Fine-tuned LLMs}} \\
Llama-3-8B-SFT       & 8B & 37.52 & 59.96 & 89.45 & 0.684 & 0.292 & 0.531 \\
Stage~I              & 8B & 27.08 & 56.22 & 89.04 & \textbf{0.957} & 0.346 & 0.621 \\
SPG (Ours)      & 8B & \textbf{37.66} & \textbf{64.18} & \textbf{90.04} & 0.827 & \textbf{0.478} & \textbf{0.634} \\
\bottomrule
\end{tabular}
\caption{Automatic evaluation on the HUPD-DCG test set (1,311 samples). All models use temperature${=}0.1$ and max\_new\_tokens${=}1{,}024$. R-1\,=\,ROUGE-1, BS\,=\,BERTScore F1. SP (Structural Pattern, 0--1), AC (Antecedent Consistency, 0--1), and CC (Claim Count ratio, 0--1.5) are surface proxies that overlap with the preference reward and are reported as auxiliary indicators only; dependency-edge accuracy (Table~\ref{tab:dea}) is our primary structural metric.}
\label{tab:main_results}
\end{table*}

\section{Experiment}
\subsection{Experimental Setup}
We evaluate on HUPD-DCG~\citep{jiang-etal-2025-hupd-dcg}, building on Llama-3-8B-instruct with LoRA on a single 80\,GB GPU\@. Baselines are chosen to span two axes, domain specialization (domain-specific vs.\ general-purpose) and parameter scale (6B--141B, dense and MoE), with exact model names and sources listed in Appendix~\ref{app:baselines}. Stage~I trains the structure-aware objective; Stage~II applies LPO using offline candidate pairs scored by PatentScore\textsubscript{rule}. Detailed in Appendix~\ref{app:hyperparams} and~\ref{app:patentscore}.

\subsection{Evaluation Metrics}
\label{sec:metrics}
We group the automatic metrics by their relation to the training signal, since
only metrics disjoint from the preference reward (\S\ref{sec:lpo}) can serve as
independent evidence.

\paragraph{Text quality.}
BLEU~\citep{papineni2002bleu} and ROUGE-1~\citep{lin2004rouge} measure lexical overlap with the reference claims, while BERTScore~\citep{zhang2020bertscore} measures semantic similarity via RoBERTa-large~\citep{liu2019robertarobustlyoptimizedbert} embeddings. The reward contains no semantic-embedding term, so BERTScore reports on an axis training never optimises.\looseness=-1

\paragraph{Structural metric (primary).}
Our central claim concerns whether the predicted claim tree matches the true
one, so we take \textbf{dependency-edge accuracy} (DEA) as primary: the
fraction of dependent claims whose predicted parent matches the gold parent in
the annotated forest. No term in the reward supervises edge accuracy.

\paragraph{Structure-aware proxies (auxiliary).}
Three surface proxies follow the conventions of prior work. \textbf{Structural Pattern} (SP, 0--1) is a weighted sum over transitional phrases, semicolon-delimited elements, and explicit parent references; \textbf{Antecedent Consistency} (AC, 0--1) is the fraction of definite references (\emph{the~X}, \emph{said~X}) resolved by a prior indefinite introduction, a missing basis being a standard ground for rejection under 35~U.S.C.~\S112(b); \textbf{Claim Count ratio} (CC, 0--1.5) is $\min(n_{\text{gen}}, n_{\text{ref}})/\max(n_{\text{gen}}, n_{\text{ref}}) \times 1.5$, penalising over- and under-generation alike. Full definitions appear in Appendix~\ref{app:patentscore}. Because these three share components with PatentScore\textsubscript{rule}, we read them as indicators rather than independent evidence. None of the rule-based metrics captures substantive legal validity, which we assess separately through blind expert evaluation.

\begin{table}[h]
\centering
\small
\setlength{\tabcolsep}{5pt}
\begin{tabular}{@{}lcccccc@{}}
\toprule
& \textbf{Overall} & \textbf{d2} & \textbf{d3} & \textbf{d4} & \textbf{d5} \\
\midrule
DEA & 0.790 & 0.925 & 0.738 & 0.604 & 0.524 \\
\bottomrule
\end{tabular}
\caption{Dependency-edge accuracy of SPG, overall and by depth ($\text{d}k$, root${=}$depth~1); no reward term supervises this metric. Depths $\geq 6$ omitted (sparse).}
\label{tab:dea}
\end{table}

\subsection{Main Results}
\label{sec:main_results}

Table~\ref{tab:main_results} summarises the automatic evaluation on the
HUPD-DCG test set.

\paragraph{The predicted topology matches the gold structure.}
Dependency-edge accuracy verifies our central structural claim directly (Table~\ref{tab:dea}): SPG recovers the correct parent for 79.0\% of dependent claims, measured against the annotated forest and on an axis the reward never supervises. Accuracy is high at shallow depths (92.5\% at depth~2) and falls with depth (52.4\% at depth~5), a gradient that anticipates the referential behaviour discussed below. No baseline appears in Table~\ref{tab:dea} because none predicts a parent: models that emit claims as flat text expose no topology to score. \looseness=-1

\paragraph{Text quality.}
Stage~I+II attains the best BLEU (37.66), ROUGE-1 (64.18), and BERTScore
(90.04) in Table~\ref{tab:main_results}. The +4.22 ROUGE-1 gain over the
supervised baseline, against a BLEU difference of only +0.14, indicates that
preference optimisation broadens coverage of technical elements rather than
polishing surface fluency. The BERTScore gain is the more informative of the
two, since no semantic-embedding term enters the reward and it therefore
cannot be attributed to reward overfitting.

\paragraph{Structural competence does not emerge from scaling or domain
pretraining.}
Every zero-shot model receives the same structured prompt specifying the
required claim format, including parent-claim references and hierarchical
scope narrowing (Appendix~\ref{app:prompts}). Yet from 6B to 141B parameters,
SP stays at or below 0.221 for all but one model while BLEU spans 12--37:
Mixtral-8$\times$22B posts the highest zero-shot ROUGE-1 (60.57) but no
structural advantage over Mistral-7B at 20$\times$ fewer parameters
(0.221 vs.\ 0.165). Fluency accumulates from local inter-token coherence and
scales with parameters; structural compliance requires satisfying relations
between non-adjacent segments, for which neither prompting nor scale supplies
an inductive bias. Domain pretraining does not substitute for one either, as
PatentGPT-J-6B and SaulLM-7B remain at SP~$\leq$~0.162 despite extensive
patent exposure. Qwen-3.5-9B is the lone high-SP zero-shot model (0.754), but
its BLEU of 19.13 is the lowest among general-purpose models, indicating
structural markers largely decoupled from the input; the contrast illustrates
that SP alone cannot separate genuine structure from its surface signature,
which is why we treat DEA as primary.

\paragraph{Referential coherence as a function of tree distance.}
The sharpest contrast is between Llama-3-8B-SFT and the full model: BLEU is
effectively tied (37.52 vs.\ 37.66) while AC differs substantially
(0.292 vs.\ 0.478). Antecedent consistency requires a definite reference
(\textit{the~X}) to resolve against an indefinite introduction (\textit{a~X})
that may sit in a structurally distant ancestor. In a flat sequence such
resolution is governed by linear token distance, whereas in a claim tree a
depth-3 claim must resolve against its depth-1 root irrespective of
intervening siblings. Because the pointer and scope gradients pass through the
shared backbone during training, rather than a parent vector being re-injected
at inference, tree distance becomes available to the decoder as an operative
factor; consistent with this, the pointer's predictions and the generated
back-references agree on 78.7\% of dependent claims. We note this agreement is
correlational and does not by itself establish that the pointer causes the
back-reference. Violation-weighted gradients then reinforce the same
discipline, assigning stronger signal to outputs with unresolved cross-claim
references than to local disfluencies.

\paragraph{Two-stage complementarity.}
Stage~I reaches the highest SP (0.957) at the cost of BLEU (27.08), exposing a
gradient competition between structural and language-modelling objectives that
single-stage training does not resolve. Stage~II restores fluency
(BLEU~37.66) while improving AC (0.346$\to$0.478) and CC
(0.621$\to$0.634). The concurrent SP decline to 0.827 is consistent with
dilution rather than structural loss: Stage~II claims are longer and
lexically more varied, which lowers the density of the surface markers SP
counts, while the referential metrics improve. This supports a
\emph{rigidity-then-flexibility} principle, in which structure is
over-constrained first and content optimised within the resulting scaffold.

%%%%%%%%%%%%%%%%%%%%%%%%%%%%%%%%%%%%%%%%%%%%%%%%%%%%%%%
%%%%%%%%%%%%%%%%%%%%%%%%%%%%%%%%%%%%%%%%%%%%%%%%%%%%%%%
%%%%%%%%%%%%%%%%%%%%%%%%%%%%%%%%%%%%%%%%%%%%%%%%%%%%%%%
%%%%%%%%%%%%%%%%%%%%%%%%%%%%%%%%%%%%%%%%%%%%%%%%%%%%%%%
\subsection{Expert Evaluation}
\label{sec:human}

Two patent practitioners, an attorney with 8 years of professional experience
and a researcher holding a PhD in patent law, independently rated 200
randomly sampled test instances on a 1--5 Likert scale along two dimensions:
\emph{Structural quality} (well-formed dependency structure, back-references,
antecedent usage) and \emph{Semantic quality} (fidelity to the inventive
concept and appropriateness of scope). Outputs were shuffled and
model-anonymised; rubrics and procedure are in Appendix~\ref{app:annotation}.

\begin{table}[h]
\centering
\small
\setlength{\tabcolsep}{4pt}
\begin{tabular}{@{}p{0.49\linewidth}ccc@{}}
\toprule
\textbf{Model} & \textbf{Struct.} & \textbf{Sem.} & \textbf{Overall} \\
\midrule
Llama-3-8B               & 2.14 & 2.87 & 2.51 \\
Llama-3-8B-SFT           & 3.62 & 3.41 & 3.52 \\
Llama-3-70B              & 2.43 & 3.58 & 3.01 \\
Full model (Ours)        & \textbf{4.21} & \textbf{3.89} & \textbf{4.05} \\
\midrule
\multicolumn{4}{@{}l}{Inter-annotator agreement: $\kappa$ = 0.67 (Cohen's)} \\
\bottomrule
\end{tabular}
\caption{Expert evaluation on 200 randomly sampled test instances (Likert 1--5). $\kappa$\,=\,Cohen's kappa.}
\label{tab:human_eval}
\end{table}

Two aspects of Table~\ref{tab:human_eval} bear on our claims. First,
Llama-3-70B outscores the 8B SFT model on semantics yet sits near the
zero-shot 8B baseline on structure, so the dissociation between fluency and
structural compliance is visible to practitioners and not an artefact of
rule-based scoring. Second, the structural ranking the experts produce
(Ours $>$ SFT $>$ 70B $>$ 8B) matches the one our automatic structure-aware
metrics produce, which mitigates the concern that those proxies reward
formatting detached from genuine hierarchical well-formedness. Agreement is
substantial ($\kappa = 0.67$).

%% file: sec6_result.tex
\section{Ablation Study}
\label{sec:ablation}

\begin{table}[h]
\centering
\small
\setlength{\tabcolsep}{3pt}
\begin{tabular}{@{}lcccccc@{}}
\toprule
\textbf{Config.} & \textbf{BLEU} & \textbf{R-1} & \textbf{BS} & \textbf{SP} & \textbf{AC} & \textbf{CC} \\
\midrule
Baseline & 37.52 & 59.96 & 89.45 & 0.684 & 0.292 & 0.531 \\
$\mathcal{L}_{\text{LM}}$ only & 35.00 & 50.91 & 89.39 & 0.953 & 0.391 & \textbf{0.798} \\
Stage~I & 27.08 & 56.22 & 89.04 & \textbf{0.957} & 0.346 & 0.621 \\
+DPO ($\omega{=}1$) & 4.07 & 16.74 & 82.56 & 0.633 & 0.153 & 0.210 \\
SPG (Ours) & \textbf{37.66} & \textbf{64.18} & \textbf{90.04} & 0.827 & \textbf{0.478} & 0.634 \\
\bottomrule
\end{tabular}
\caption{Ablation on HUPD-DCG. Baseline trains on plain claims; $\mathcal{L}_{\text{LM}}$ only and subsequent rows train on structure-tagged data.}
\label{tab:ablation}
\end{table}

\paragraph{Structural tags fix formatting, not depth.}
Table~\ref{tab:ablation} isolates each component. The Baseline, trained on plain claims, leads the single-objective configurations on BLEU (37.52) but trails on structure (SP=0.684, AC=0.292). Annotating claim boundaries alone raises SP to 0.953 with no pointer supervision, so surface formatting is largely a matter of data representation: once boundaries are visible in the sequence, the autoregressive objective reproduces the conventions SP counts, and AC improves to 0.391 within each delimited unit. Depth is what annotation does not buy. R-1 falls to 50.91 while CC reaches its highest value (0.798), a combination that is only apparently contradictory: the model emits roughly the right \emph{number} of claims but places nearly all of them at depth~1, so the count ratio is satisfied by flat trees that leave most reference content uncovered.

\paragraph{Effect of pointer supervision.}
Stage~I ($\gamma{=}1.0$) recovers R-1 to 56.22 while holding SP at 0.957. Committing to a parent at each \texttt{<dep>} boundary sustains generation past the first level and anchors subsequent tokens to the selected claim, yielding deeper trees whose content tracks each parent. AC falls in the process (0.391$\to$0.346), which we read as a harder task rather than a worse model: resolution now spans multiple depths, and depth is precisely where accuracy degrades (Table~\ref{tab:dea}). A $\gamma$ sweep appears in Appendix~\ref{app:loss_ablation}.

\paragraph{Preference optimisation and mode collapse.}
Standard DPO ($\omega{=}1$) applied to the Stage~I checkpoint collapses. The failure is catastrophic rather than merely repetitive, since a truncated independent claim invalidates every dependent claim beneath it, and uniform gradients sharpen this fragility by scoring a rejected sample that omits one semicolon exactly as one that omits an entire subtree. Weighting each pair by $\omega(y_l) = 1 - \text{PS}(y_l)/\text{PS}_{\text{max}}$ restores stable convergence, and a magnitude-matched control isolates why: a uniform objective whose gradient norm is rescaled to that of LPO reaches a valid rate of 71.4\% (the fraction of parseable, complete claim forests; Appendix~\ref{app:hyperparams}), far above standard DPO at 31.6\% but short of LPO at 84.2\%. Severity weighting accounts for the remaining 71.4$\to$84.2 gap, which step size alone leaves unexplained, though on a single run with a large margin (see Limitations). The full model correspondingly restores BLEU to 37.66 and attains the highest AC of any configuration (0.478), exceeding both Stage~I and $\mathcal{L}_{\text{LM}}$ only, which suggests the contrastive signal supplies referential discipline across deep chains that neither objective provides alone.

%% file: sec7_conclusion.tex
\section{Discussion}

Our three findings answer one question at three levels of the system: what supplies the inductive bias that a next-token objective does not? We state each as a prescription, since each should transfer to generation tasks whose outputs carry enforceable structure.

\paragraph{Finding 1: Scale is not a substitute for structural supervision.}
Structural compliance stayed flat from 6B to 141B parameters while fluency rose steadily, and domain pretraining moved it no further (\S\ref{sec:main_results}). The implication is a design order rather than a ranking of models: when a task requires relations between non-adjacent segments, an explicit structural mechanism should precede a larger backbone, because the two are not interchangeable. Our Qwen-3.5-9B case adds that even a structure-sensitive surface metric can be satisfied by markers detached from content, so topology must be scored against a reference tree rather than inferred from formatting.

\paragraph{Finding 2: Preference optimisation over structured outputs needs
severity-aware weighting.}
Deficiencies in structured generation are not exchangeable: a missing semicolon and a missing subtree are one preference pair each under uniform weighting, yet only the latter invalidates everything that references it. The collapse of uniform DPO, and its partial recovery under a magnitude-matched control, indicate that the weighting function does work that gradient scale alone does not (\S\ref{sec:ablation}). Where output parts depend on other parts, we would treat the severity function as a first-class design choice, and report a magnitude-matched baseline whenever a reweighted objective is claimed to help, since otherwise its effect is not separable from a smaller effective step size. Our evidence rests on a single run and should be read accordingly.

\paragraph{Finding 3: Structure and content are better separated in time than balanced by coefficients.}
Weighting the structural and language-modelling losses against each other within one stage traded one for the other in every configuration we tried (Appendix~\ref{app:loss_ablation}), whereas building the structural scaffold first and optimising content within it improved both. The recipe for objectives competing over the same representations is therefore to over-constrain the harder-to-recover property first, then relax toward fluency, and to expect surface-level structural metrics to dilute even as substantive ones improve, since a metric that falls for the right reason is easily mistaken for a regression.

\section{Conclusion}

We studied the generation of tree-structured outputs whose nodes must remain semantically consistent with their ancestors, using patent claims as a testbed because the hierarchy there is annotated, statutory, and automatically checkable. SPG predicts topology inside the autoregressive pass, so that a pointer head's gradients reshape the same representations that emit the tokens; a depth-adaptive margin constrains the parent-child relation those gradients bring into existence, and violation-weighted preference optimisation supplies the negative signal that granted-patent corpora withhold.

On HUPD-DCG, an 8B model equipped with SPG recovers 79.0\% of gold parent links, a quantity its reward never supervises, and improves referential consistency over an equal-scale supervised baseline while matching it on fluency; it also exceeds much larger zero-shot models on our structure-aware metrics, though those comparisons are reference points rather than head-to-head contrasts, and expert ratings order the systems as the automatic structural metrics do. Beyond patents, hierarchical generation appears to benefit less from scale than from placing structural decisions inside the decoder, weighting preferences by severity, and separating structure and content in time.

%% file: sec8_limitation_ethics.tex
\section*{Limitations}
This work targets U.S. patent claims filed under the USPTO framework and presupposes a single-parent dependency structure. However, patent law governs claim dependency formats differently across jurisdictions. The EPO, pursuant to EPC Rule 43(4), permits multiple dependent claims that simultaneously reference several antecedent claims, yielding a directed acyclic graph topology rather than a forest. The JPO and KIPO similarly allow multiple dependencies, albeit with divergent prosecution guidelines regarding permissible depth and combinatorial scope of such dependencies. Extending the framework to multi-jurisdictional dependency structures necessitates not only a redefinition of the topological formalism but also a re-establishment of jurisdiction-specific scope narrowing criteria, both of which fall beyond the scope of the present study.

Evaluation in this work is confined to monolingual English patents. Multilingual claim drafting in PCT applications demands not merely one-to-one cross-lingual terminological correspondence but simultaneous accommodation of jurisdiction-specific claim drafting conventions that differ structurally. For instance, the varying prevalence of Jepson-type claims and the divergent interpretive scope of means-plus-function claims across patent offices. These considerations constitute an independent research problem orthogonal to the hierarchical structural constraints addressed herein.

Our experiments also carry methodological limitations. Results are reported from single runs without multiple seeds, confidence intervals, or significance testing, so the reported gaps, including the LPO-versus-DPO comparison, should be read as indicative rather than statistically established. Moreover, our automatic structural metrics are rule-based and do not fully capture substantive legal validity; the system is intended to assist patent professionals, and its outputs require expert review prior to filing. The 141B comparison, in particular, is not a head-to-head contrast but a zero-shot reference point supporting Finding~1, since those baselines are not fine-tuned on the task.

Finally, this work concentrates on the generation of claim text and dependency structure. The broader patent drafting workflow involves document-level strategic reasoning, including upper conceptualization of the invention, novelty and inventive step argumentation against prior art, and claim scope calibration informed by prosecution history estoppel. Such prosecution-strategy-level decisions reside at a fundamentally different abstraction layer from claim-level generation and lie outside the scope of this study.

\section*{Ethical Considerations}
During the preparation of this work, the author(s) utilized generative AI to refine linguistic clarity and support the creation of certain diagrams. The author(s) carefully reviewed all outputs and maintain full responsibility for the intellectual content and originality of the final paper. Upon acceptance, we will release the code, the trained models, the LPO preference pairs, and the evaluation scripts to support reproducibility.

%% file: sec8_appendix.tex
\appendix

\section{Baseline Models}
\label{app:baselines}
Table~\ref{tab:baselines} lists the official name, size, and source of each baseline used in Table~\ref{tab:main_results}, spanning domain-specific and general-purpose models across dense and MoE architectures from 6B to 141B parameters. All fine-tuned variants (SFT, Stage~I, SPG) share \texttt{Meta-Llama-3-8B-Instruct} as the backbone, isolating the effect of our method from scale and domain pretraining.

\begin{table}[h]
\centering
\footnotesize
\setlength{\tabcolsep}{4pt}
\begin{tabular}{@{}p{0.28\columnwidth}l p{0.44\columnwidth}@{}}
\toprule
Model & Params & Source \\
\midrule
PatentGPT-J-6B       & 6B   & \texttt{patent/PatentGPT-J-6B} \\
SaulLM-7B            & 7B   & \texttt{Equall/\allowbreak Saul-7B-\allowbreak Instruct-v1} \\
Mistral-7B           & 7B   & \texttt{mistralai/\allowbreak Mistral-7B-\allowbreak Instruct-v0.3} \\
Llama-3-8B           & 8B   & \texttt{meta-llama/\allowbreak Meta-Llama-3-\allowbreak 8B-Instruct} \\
Qwen-3.5-9B          & 9B   & \texttt{Qwen/Qwen3.5-9B} \\
Llama-3-70B          & 70B  & \texttt{meta-llama/\allowbreak Meta-Llama-3-\allowbreak 70B-Instruct} \\
Mixtral-8$\times$7B  & 47B  & \texttt{mistralai/\allowbreak Mixtral-8x7B-\allowbreak Instruct-v0.1} \\
Mixtral-8$\times$22B & 141B & \texttt{mistralai/\allowbreak Mixtral-8x22B-\allowbreak Instruct-v0.1} \\
\bottomrule
\end{tabular}
\caption{Baseline models, sizes, and sources. Fine-tuned variants (SFT, Stage~I, SPG) all use \texttt{Meta-Llama-3-8B-Instruct} as the backbone.}
\label{tab:baselines}
\end{table}
% ====================================================================

\section{Prompts}
\label{app:prompts}
All zero-shot baselines are evaluated with the same structured prompt, which specifies the required claim format (independent vs.\ dependent claims, explicit parent-claim references, and hierarchical scope narrowing) followed by the patent description. The full prompt text is reproduced below.

\begin{quote}
\ttfamily
% TODO: paste the exact zero-shot prompt string here from the codebase
\end{quote}

% ====================================================================
\section{Dataset Statistics}
\label{app:dataset}

HUPD-DCG~\citep{jiang-etal-2025-hupd-dcg} comprises
9,555 granted U.S.\ patent documents filed in 2017,
filtered from the Harvard USPTO Patent
Dataset~\citep{suzgun2023hupd} to include only
descriptions shorter than 8,000 tokens.
Table~\ref{tab:data_split} summarizes the train/test
partition, and Table~\ref{tab:data_detail} reports
detailed statistics.
Train and test distributions are closely aligned across
all metrics, confirming the temporal split does not
introduce distributional shift.

\begin{table}[h]
\centering
\small
\begin{tabular}{@{}p{0.55\linewidth}p{0.15\linewidth}p{0.15\linewidth}@{}}
\toprule
 & \textbf{Train} & \textbf{Test} \\
\midrule
Documents            & 8,244  & 1,311  \\
Avg.\ claims          & 14.2   & 13.6   \\
Avg.\ independent     &  2.3   &  2.2   \\
Avg.\ dependent       & 11.9   & 11.4   \\
Dep.\ ratio (\%)      & 84.1   & 84.1   \\
Avg.\ max depth       &  2.4   &  2.3   \\
\bottomrule
\end{tabular}
\caption{Train/test split overview. The dependent
claim ratio and average tree depth are nearly
identical across splits.}
\label{tab:data_split}
\end{table}

\begin{table}[t]
\centering
\small
\setlength{\tabcolsep}{4pt}
\begin{tabular}{lrrrr}
\toprule
 & \textbf{Mean} & \textbf{Med.}
 & \textbf{Min} & \textbf{Max} \\
\midrule
\multicolumn{5}{l}{\emph{Train set
  (8,244 documents)}} \\[2pt]
Claims / doc
  & 14.2  &  15  &   0 &    75 \\
Independent / doc
  &  2.3  &   2  &   0 &    31 \\
Dependent / doc
  & 11.9  &  12  &   0 &    72 \\
Max tree depth
  &  2.4  &   2  &   0 &    14 \\
Claim tokens
  &  984  & 920  &  50 & 6,894 \\
Desc.\ tokens
  & 5,798 & 5,960 & 855 & 9,355 \\
\midrule
\multicolumn{5}{l}{\emph{Test set
  (1,311 documents)}} \\[2pt]
Claims / doc
  & 13.6  &  14  &   0 &    51 \\
Independent / doc
  &  2.2  &   2  &   0 &    17 \\
Dependent / doc
  & 11.4  &  12  &   0 &    47 \\
Max tree depth
  &  2.3  &   2  &   0 &    16 \\
Claim tokens
  &  976  & 893  &   7 & 4,406 \\
Desc.\ tokens
  & 5,757 & 5,946 & 1,298 & 9,089 \\
\bottomrule
\end{tabular}
\caption{Detailed statistics. All token counts use the
Qwen-2.5 tokenizer. Description lengths are below
${\sim}$9.4K tokens by construction, and claim sets
average ${\sim}$1K tokens.}
\label{tab:data_detail}
\end{table}

\begin{figure}[h]
\centering
\includegraphics[width=\columnwidth]{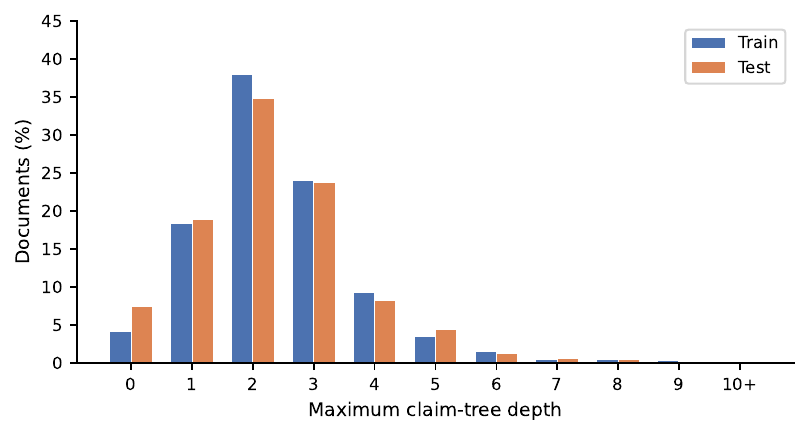}
\caption{Distribution of maximum claim-tree depth.
Depths 1--3 account for over 80\% of documents in
both splits.}
\label{fig:depth_dist}
\end{figure}

Figure~\ref{fig:depth_dist} shows the distribution of
maximum claim-tree depths.
The majority of patents (80\% train, 77\% test) have a
maximum depth between 1 and 3, indicating relatively
shallow dependency structures.
A long tail extends to depth 14--16, motivating our
depth-adaptive designs
(Eqs.~\ref{eq:scope}).

\paragraph{Dependency parsing coverage.} 
Claim dependencies are extracted via pattern matching on referencing expressions (e.g., ``The system of claim~1,'' ``A device according to claim~3''). Of the 116,813 training claims, 98,236 (84.1\%) are identified as dependent and 17,839 (15.3\%) as independent.

%%%%%%%%%%%%%%%%%%%%%%%%%%%%%%%%%%%%%%%%%%%%%%%%%
%%%%%%%%%%%%%%%%%%%%%%%%%%%%%%%%%%%%%%%%%%%%%%%%%
%%%%%%%%%%%%%%%%%%%%%%%%%%%%%%%%%%%%%%%%%%%%%%%%%

\section{Hyperparameter Details}
\label{app:hyperparams}

\subsection{Model Configuration}

We use \texttt{meta-llama/Meta-Llama-3-8B-Instruct} as the base model. Six special tokens are added to the vocabulary to encode claim structure: \texttt{<ind>} and \texttt{</ind>} delimit independent claims; \texttt{<dep>} and \texttt{</dep>} delimit dependent claims; \texttt{<sep>} separates trees within the forest; \texttt{<eot>} marks the end of the claim set. The embedding and language model head matrices are resized accordingly, with new embeddings initialised from the mean of existing embeddings.

\subsection{Stage~I: Structure-Aware Fine-Tuning}

Table~\ref{tab:hparams_s1} lists all Stage~I hyperparameters. We apply LoRA~\citep{hu2022lora} to the query and value projection matrices. The maximum context length is 8{,}192 tokens; inputs exceeding this limit are truncated from the left. Training uses mixed-precision (bfloat16) and gradient checkpointing to fit within a single 80\,GB GPU\@.

\begin{table}[h]
\centering
\small
\caption{Stage~I hyperparameters.}
\label{tab:hparams_s1}
\begin{tabular}{@{}p{0.55\linewidth}p{0.35\linewidth}@{}}
\toprule
\textbf{Parameter} & \textbf{Value} \\
\midrule
LoRA rank ($r$)          & 96 \\
LoRA alpha ($\alpha$)    & 32 \\
LoRA dropout             & 0.05 \\
LoRA target modules      & all linear \\
Learning rate            & $2 \times 10^{-4}$ \\
LR scheduler             & Cosine \\
Warmup ratio             & 0.05 \\
Weight decay             & 0.01 \\
Batch size (per device)  & 1 \\
Gradient accumulation    & 8 \\
Effective batch size     & 8 \\
Epochs                   & 5 \\
Max context length       & 8{,}192 \\
Structure loss weight ($\gamma$) & 1.0 \\
Scope loss weight ($\eta$)       & 0.01 \\
Scope base radius ($\rho$)       & 5.0 \\
Scope decay factor ($\lambda$)   & 0.85 \\
Precision                & bfloat16 \\
\bottomrule
\end{tabular}
\end{table}

\subsection{Stage~II: LPO-Based Preference Optimisation}

\paragraph{Offline preference pair construction.}
Starting from the Stage~I checkpoint, we generate two
candidate claim sets per training sample at temperatures
0.3 and 1.2 (\texttt{max\_new\_tokens}$\,{=}\,$768).
Each candidate is scored by
PatentScore\textsubscript{rule}; the ground-truth claim
serves as the chosen response, and the lower-scoring
candidate as the rejected response.
Up to 2{,}000 training samples are used, processed
in batches of 12.

\paragraph{LPO training.}
Table~\ref{tab:hparams_s2} lists the Stage~II
hyperparameters.
The LoRA adapter from Stage~I is loaded and a second
LoRA adapter is initialised for LPO training.
Stage~I special-token embeddings are restored before
training begins.

\begin{table}[h]
\centering
\small
\caption{Stage~II hyperparameters.}
\label{tab:hparams_s2}
\begin{tabular}{@{}p{0.55\linewidth}p{0.35\linewidth}@{}}
\toprule
\textbf{Parameter} & \textbf{Value} \\
\midrule
LoRA rank ($r$)           & 8 \\
LoRA alpha ($\alpha$)     & 16 \\
LoRA dropout              & 0.05 \\
LPO $\beta$               & 0.1 \\
Learning rate             & $5 \times 10^{-6}$ \\
LR scheduler              & Cosine \\
Warmup ratio              & 0.1 \\
Batch size (per device)   & 7 \\
Gradient accumulation     & 4 \\
Epochs                    & 3 \\
Max sequence length       & 2{,}048 \\
Pair temperatures         & \{0.3, 1.2\} \\
Max pairs generated       & 2{,}000 \\
Best-vs-worst margin ($\kappa$) & 0.3 \\
Gt-vs-best threshold ($\delta$) & 0.1 \\
$\text{PS}_{\max}$        & 4.5 \\
Precision                 & bfloat16 \\
\bottomrule
\end{tabular}
\end{table}

\subsection{Evaluation Settings}

All models are evaluated with greedy-like decoding
(\texttt{temperature}$\,{=}\,$0.1,
\texttt{do\_sample}$\,{=}\,$True).
The maximum generation length is 1{,}024 tokens.
Prompts exceeding $8{,}192 - 1{,}024 = 7{,}168$ tokens
are truncated.
BERTScore is computed using \texttt{roberta-large}.
Evaluation proceeds in batches of 12 with left-padding.

\paragraph{Valid rate.}
Beyond the metrics of \S\ref{sec:metrics}, we report a \emph{valid rate} in the
preference-objective comparison (\S\ref{sec:ablation}): the fraction of test
outputs that parse into a complete claim forest. An output counts as valid when
(i) its structural delimiters are balanced and it terminates with \texttt{<eot>}
within the generation budget, (ii) it contains at least one well-formed
independent claim, and (iii) every dependent claim resolves to a parent that
exists and precedes it in the serialisation. Outputs that are truncated,
degenerate, or contain dangling references are counted as invalid. The measure
is deliberately coarse: it detects the catastrophic failures that mode collapse
produces, and says nothing about claim quality.

\begin{table}[h]
\centering
\small
\caption{Evaluation settings.}
\label{tab:eval_settings}
\begin{tabular}{@{}p{0.55\linewidth}p{0.35\linewidth}@{}}
\toprule
\textbf{Parameter} & \textbf{Value} \\
\midrule
Temperature              & 0.1 \\
\texttt{do\_sample}      & True \\
\texttt{max\_new\_tokens} & 1{,}024 \\
Max context              & 8{,}192 \\
Prompt truncation limit  & 7{,}168 \\
Evaluation batch size    & 4 \\
BERTScore model          & \texttt{roberta-large} \\
\bottomrule
\end{tabular}
\end{table}

%%%%%%%%%%%%%%%%%%%%%%%%%%%%%%%%%%%%%%%%%%%%%%%%%%%%%
%%%%%%%%%%%%%%%%%%%%%%%%%%%%%%%%%%%%%%%%%%%%%%%%%%%%%
\section{Notation}

Table~\ref{tab:notation} summarises the key symbols used throughout the paper. Calligraphic letters denote sets and structures, bold lowercase letters denote vectors or individual claims, and bold uppercase letters denote matrices.

\begin{table}[H]
\centering
\small
\caption{Main notation.}
\label{tab:notation}
\begin{tabular}{@{}p{0.45\linewidth}p{0.45\linewidth}@{}}
\toprule
\textbf{Symbol} & \textbf{Definition} \\
\midrule
$\cD$ & patent description \\
$\mathcal{C}$ & claim set \\
$\cF=\{\cT_1,\dots,\cT_K\}$ & claim dependency forest \\
$\cT_k$ & $k$-th dependency tree \\
$\cV=\{\mbv_1,\dots,\mbv_N\}$ & claims \\
$\cE$ & dependency edges \\
$\mbv_i$ & $i$-th claim \\
$\mbv_k^{\mathrm{root}}$ & root claim of $\cT_k$ \\
$\mathcal{S}$ & forest serialization \\
$\mathcal{S}_k$ & tree serialization \\
$\mathcal{D}_k$ & DFS order of non-root claims \\
$\mathcal{A}(i)$ & candidate parents of $\mbv_i$ \\
$\mathbf{h}_{\texttt{<dep>}}^{(i)}\!\in\!\mathbb{R}^d$ & delimiter hidden vector \\
$\mathbf{h}_{\texttt{end}}^{(j)}\!\in\!\mathbb{R}^d$ & claim-end hidden vector \\
$W_{\mathrm{dep}}\!\in\!\mathbb{R}^{d\times d}$ & parent-selection matrix \\
$\mathbf{z}_i\!\in\!\mathbb{R}^d$ & claim representation \\
$\mathrm{pa}(i)$ & parent index of $\mbv_i$ \\
$d(\mbv_i)$ & depth of $\mbv_i$ \\
$d$ & hidden dimension \\
\bottomrule
\end{tabular}
\end{table}

%%%%%%%%%%%%%%%%%%%%%%%%%%%%%%%%%%%%%%%%%%%%%%%%%%
%%%%%%%%%%%%%%%%%%%%%%%%%%%%%%%%%%%%%%%%%%%%%%%%%%
%%%%%%%%%%%%%%%%%%%%%%%%%%%%%%%%%%%%%%%%%%%%%%%%%%

%%%%%%%%%%%%%%%%%%%%%%%%%%%%%%%%%%%%%%%%%%%%%%%%%%
%%%%%%%%%%%%%%%%%%%%%%%%%%%%%%%%%%%%%%%%%%%%%%%%%%
%%%%%%%%%%%%%%%%%%%%%%%%%%%%%%%%%%%%%%%%%%%%%%%%%%

\section{PatentScore\textsubscript{rule}: Implementation Details}
\label{app:patentscore}

\subsection{Motivation}

Legal Preference Optimisation (§\ref{sec:lpo}) requires a scoring function $\text{PS}(y) \in [0, 4.5]$ that ranks generated claim sets by structural and content quality so that preference pairs can be constructed without human annotation. The function must satisfy three requirements: (i)~\emph{deterministic}, so that the same candidate always receives the same score and training is reproducible; (ii)~\emph{format-neutral}, so that candidates produced with structural markup tokens (\texttt{<ind>}, \texttt{<dep>}, \texttt{<sep>}) are scored on the same basis as plain-text outputs; and (iii)~\emph{cost-free}, so that scoring thousands of candidates during preference construction does not introduce API costs or GPU overhead.

We design \textbf{PatentScore\textsubscript{rule}} to meet these requirements. The function evaluates structural completeness, claim count fidelity, and lexical overlap with the reference claim set through five rule-based sub-metrics, each capturing a dimension of the original PatentScore~\citep{yoo-etal-2025-patentscore}. Table~\ref{tab:ps_mapping} summarises the mapping.

\begin{table}[t]
\centering
\small
\begin{tabular}{@{}lp{1.4cm}cc@{}}
\toprule
\textbf{Sub-Metric} & \textbf{Original} & \textbf{Max} & \textbf{Type} \\
\midrule
Claim Count Match &
  $M_\text{VU}$ & 1.5 & Rule \\
Structural Pattern &
  $M_\text{CS}$, $M_\text{CP}$ & 1.0 & Rule \\
Antecedent Consistency &
  $M_\text{AB}$, $M_\text{ER}$ & 1.0 & Rule \\
Length Ratio &
  $M_\text{AS}$ & 0.5 & Rule \\
Content Overlap &
  $M_\text{BS}$ & 0.5 & Rule \\
\bottomrule
\end{tabular}
\caption{Mapping between PatentScore\textsubscript{rule}
sub-metrics and the original PatentScore dimensions.}
\label{tab:ps_mapping}
\end{table}

\subsection{Format-Neutral Preprocessing}

Before scoring, every candidate and reference claim set
undergoes identical preprocessing:

\begin{enumerate}
  \item Remove all structural markup tokens
    (\texttt{<ind>}, \texttt{</ind>}, \texttt{<dep>},
     \texttt{</dep>}, \texttt{<sep>}, \texttt{<eot>}).
  \item Remove model-specific tokens
    (e.g., \texttt{<|eot\_id|>}).
  \item Normalise whitespace (collapse multiple spaces,
    strip leading/trailing whitespace).
\end{enumerate}

\noindent This ensures that the scoring function is
agnostic to whether the candidate was generated with or
without structural markup, a necessary condition for
constructing preference pairs from heterogeneous decoding
configurations (e.g., different sampling temperatures).

\subsection{Sub-Metric Definitions}

PatentScore\textsubscript{rule} comprises five sub-metrics.
Let $\hat{y}$ denote the preprocessed candidate and
$y$ the preprocessed reference claim set.

\paragraph{(1) Claim Count Match (max 1.5).} This sub-metric captures the validity and uniqueness dimension ($M_\text{VU}$) by measuring whether the number of generated claims matches the reference. Claims are counted by detecting sequential numbering patterns (e.g., ``\texttt{1.}'', ``\texttt{2.}'') at sentence boundaries and back-reference phrases (e.g., ``The method of claim~$k$''). Let $\hat{n}$ and $n_\text{gt}$ denote the candidate and reference claim counts:

\begin{equation}
  S_\text{count} =
    \frac{\min(\hat{n},\, n_\text{gt})}
         {\max(\hat{n},\, n_\text{gt})}
    \times 1.5\,.
  \label{eq:count}
\end{equation}

\paragraph{(2) Structural Pattern (max 1.0).} Patent claims follow a canonical structure: a preamble with a transitional phrase, body elements separated by semicolons, and dependent claims with explicit back-references. This sub-metric captures claim structure ($M_\text{CS}$) and punctuation ($M_\text{CP}$):

\begin{equation}
\label{eq:struct}
  S_\text{struct} =
    \alpha_\text{pre} + \alpha_\text{semi}
    + \alpha_\text{dep}\,,
\end{equation}

where:
$\alpha_\text{pre} \in \{0, 0.4\}$ indicates whether the first claim contains a preamble with a colon-terminated transitional phrase (e.g., ``comprising:'', ``consisting of:''); $\alpha_\text{semi} \in [0, 0.3]$ is the fraction of claims containing at least one semicolon, scaled by $0.3$; $\alpha_\text{dep} \in [0, 0.3]$ is the fraction of non-first claims containing a proper back-reference (e.g., ``The \ldots{} of claim'', ``according to claim''), scaled by $0.3$. This sub-metric checks only for the \emph{presence} of a transitional phrase, not whether the specific phrase used is legally appropriate for the claim in question; substantive correctness is instead assessed through the expert evaluation(Table~\ref{tab:human_eval}).

\paragraph{(3) Antecedent Consistency (max 1.0).}
Following MPEP~\S2173.05(e) and WIPO guidelines, every definite reference (``the~$X$'', ``said~$X$'') must have a prior indefinite introduction (``a~$X$'', ``an~$X$''). This sub-metric jointly captures antecedent basis ($M_\text{AB}$) and element referencing ($M_\text{ER}$): 

\begin{equation}    
  S_\text{ante} =
    \frac{|\{x \in \mathcal{D} \mid
      \exists\;\text{antecedent for } x\}|}
         {|\mathcal{D}|}\,,
  \label{eq:ante}
\end{equation}

where $\mathcal{D}$ is the set of definite noun-phrase references extracted from $\hat{y}$ via pattern matching. For each reference, we check whether its head noun (or bigram) appears in the set of indefinite introductions occurring earlier in the text. References of the form ``claim~$k$'' are automatically resolved, as they refer to other claims rather than technical entities. If $|\mathcal{D}| = 0$, the score defaults to $0$.

\paragraph{(4) Length Ratio (max 0.5).}
Excessively short or verbose claims indicate scope issues. This sub-metric serves as a proxy for the ambiguous scope dimension ($M_\text{AS}$):

\begin{equation}
  S_\text{len} =
    \frac{\min(|\hat{y}|_w,\, |y|_w)}
         {\max(|\hat{y}|_w,\, |y|_w)}
    \times 0.5\,,
  \label{eq:len}
\end{equation}
where $|\cdot|_w$ denotes the whitespace-tokenised word count.

\paragraph{(5) Content Overlap (max 0.5).}
Unigram recall against the reference approximates semantic similarity ($M_\text{BS}$):

\begin{equation}
  S_\text{overlap} =
    \frac{|\mathcal{W}(\hat{y}) \cap \mathcal{W}(y)|}
         {|\mathcal{W}(y)|}
    \times 0.5\,,
  \label{eq:overlap}
\end{equation}

where $\mathcal{W}(\cdot)$ extracts the set of lowercased whitespace-tokenised unigrams.

\subsection{Composite Score}

The composite score is the sum of all sub-metrics:
\begin{equation}
  \text{PS}_\text{rule}(y)
    = S_\text{count} + S_\text{struct} + S_\text{ante}
    + S_\text{len} + S_\text{overlap}\,,
  \label{eq:composite}
\end{equation}
with a theoretical maximum of $4.5$. This score is used directly in the preference construction procedure (§\ref{sec:lpo}): candidates are ranked by $\text{PS}_\text{rule}$, and the violation weight $\omega(y_l) = 1 - \text{PS}(y_l)/\text{PS}_\text{max}$ modulates the contrastive gradient accordingly.

\subsection{Score Calibration}

Evaluating ground-truth claims against themselves yields a score of 85.49 on the normalised 0--100 scale rather than a perfect 100, because Antecedent Consistency captures only \emph{explicit} indefinite introductions within the claim text. Human patent drafters frequently introduce entities implicitly through the description context (e.g., referring to ``the processor'' without a prior ``a processor'' within the claims), which the rule-based detector does not resolve. Structural Pattern also scores below 1.0 for ground truth (0.926) because a small fraction of real patents deviate from the canonical semicolon-delimited formatting convention. These ceiling effects are consistent across all candidates scored during preference construction and do not affect the relative ranking of preference pairs.

%%%%%%%%%%%%%%%%%%%%%%%%%%%%%%%%%%%%%%%%%%%%%%%%%%
%%%%%%%%%%%%%%%%%%%%%%%%%%%%%%%%%%%%%%%%%%%%%%%%%%
\section{Loss Component Analysis}
\label{app:loss_ablation}

\begin{table}[t]
\centering
\small
\setlength{\tabcolsep}{4pt}
\begin{tabular}{@{}lcccccc@{}}
\toprule
\textbf{Config.} & \textbf{BL} & \textbf{R-1} & \textbf{BS} & \textbf{SP} & \textbf{AC} & \textbf{CC} \\
\midrule
LLM only            & 35.00 & 50.91 & 89.39 & .953 & .391 & .798 \\
+$\mathcal{L}_{\text{scope}}$ ($\eta{=}.01$) & 34.25 & 51.07 & 89.35 & .953 & .390 & .795 \\
+$\mathcal{L}_{\text{str}}$ ($\gamma{=}0.3$) & 29.32 & 59.10 & 89.57 & .959 & .369 & .724 \\
+$\mathcal{L}_{\text{str}}$ ($\gamma{=}0.5$) & 28.14 & 57.67 & 89.29 & .957 & .368 & .677 \\
+$\mathcal{L}_{\text{str}}$ ($\gamma{=}1.0$) & 27.08 & 56.22 & 89.04 & .957 & .346 & .621 \\
\bottomrule
\end{tabular}
\caption{Stage~I loss ablation with identity-initialized $W_{\text{dep}}$. The scope radius is fixed at $\rho{=}5.0$ throughout; rows with $\mathcal{L}_{\text{str}}$ use $\eta{=}0.01$. Increasing $\gamma$ improves structural compliance (SP) while reducing fluency (BL${=}$BLEU).}
\label{tab:loss_ablation}
\end{table}

Table~\ref{tab:loss_ablation} examines the individual loss components of Stage~I by varying the structure loss weight $\gamma$, with the scope term held fixed at $\rho{=}5.0$ and $\eta{=}0.01$.

\paragraph{Structural tags drive formatting; pointer supervision drives completeness.}
With $\mathcal{L}_{\text{LM}}$ alone ($\gamma{=}0$), SP already reaches 0.953, establishing that surface-level claim formatting is predominantly a function of data representation: once \texttt{<ind>}/\texttt{<dep>} tags make claim boundaries explicit in the training sequence, the autoregressive objective learns to reproduce transitional phrases, semicolons, and back-references with high fidelity. The role of pointer supervision ($\gamma > 0$) is therefore not to improve formatting but to address a deeper structural deficiency. At $\gamma{=}0$, R-1 is only 50.91 despite SP=0.953, meaning the model produces well-formatted but shallow claim sets that omit the majority of dependent claims present in the reference. Introducing $\gamma{=}0.3$ raises R-1 to 59.10 (+8.19), the largest single-component gain in the table, because the pointer network forces the decoder to commit to a parent at each \texttt{<dep>} boundary, preventing the model from defaulting to early termination after a few independent claims.

\paragraph{Diminishing returns at higher $\gamma$.}
Beyond $\gamma{=}0.3$, R-1 decreases monotonically (59.10 $\to$ 57.67 $\to$ 56.22) while SP remains saturated at 0.957. This pattern reveals a gradient competition between the structure loss and the language modelling objective: at higher $\gamma$, the pointer supervision signal increasingly dominates the gradient, directing model capacity toward producing topologically correct trees at the expense of lexical precision within each claim. BLEU, which is sensitive to n-gram overlap, reflects this trade-off most acutely (29.32 $\to$ 27.08). We select $\gamma{=}1.0$ for Stage~I not because it maximises any single metric, but because it provides the strongest structural constraint as a foundation for Stage~II: preference optimisation subsequently recovers text quality from the $\gamma{=}1.0$ checkpoint more effectively than from weaker-constrained alternatives, as the rigid structural scaffolding prevents the preference signal from degrading claim topology during content optimisation (Table~\ref{tab:ablation}).

\paragraph{AC decreases reflect increased task difficulty, not degradation.}
AC drops from 0.391 ($\gamma{=}0$) to 0.346 ($\gamma{=}1.0$), a trend that superficially suggests pointer supervision harms referential consistency. The opposite interpretation is correct. At $\gamma{=}0$, the model generates predominantly independent claims where antecedent references resolve within a single clause, making the AC task trivial. As $\gamma$ increases, the model produces progressively deeper dependency trees where a definite reference (``the processor'') in a depth-3 claim may need to resolve against an indefinite introduction (``a processor'') located two tree levels above. The absolute AC decrease thus masks a substantial increase in the complexity of the referential task the model is attempting. This interpretation is confirmed by the R-1 trajectory: higher $\gamma$ produces more complete claim sets (higher R-1 at $\gamma{=}0.3$) with deeper trees, and it is precisely these deeper trees that make cross-claim antecedent resolution harder.

\paragraph{Scope loss requires hierarchical context.}
Adding $\mathcal{L}_{\text{scope}}$ alone ($\gamma{=}0$, $\eta{=}0.01$) produces no meaningful change (BLEU 35.00$\to$34.25, R-1 50.91$\to$51.07). This null result is structurally informative: the depth-adaptive margin constraint (Eq.~4) penalises representational distance between parent and child claims, but without pointer supervision the model generates few dependent claims and thus few parent-child pairs for the loss to act upon. The scope loss is therefore not independently ineffective but rather conditionally dependent on the hierarchical structure that $\mathcal{L}_{\text{struct}}$ creates, motivating their joint inclusion in the Stage~I objective.

\paragraph{CC reveals a generation length trade-off.}
CC decreases monotonically from 0.798 ($\gamma{=}0$) to 0.621 ($\gamma{=}1.0$). At $\gamma{=}0$, the model generates short, shallow claim sets whose claim count often happens to match the reference count by coincidence: with fewer claims generated, the numerator and denominator of the CC ratio remain close. As $\gamma$ increases, the model generates deeper trees with more dependent claims, but the fixed 1,024-token generation budget truncates longer outputs before all reference claims can be produced. The CC decrease is therefore attributable to the generation length ceiling rather than to a deficiency in the pointer mechanism itself, and is partially recovered by Stage~II (CC=0.634, Table~\ref{tab:ablation}), which produces more token-efficient claims through preference-guided content optimization. 

%%%%%%%%%%%%%%%%%%%%%%%%%%%%%%%%%%%%%%%%%%%%%%%%%%%%%%%%
%%%%%%%%%%%%%%%%%%%%%%%%%%%%%%%%%%%%%%%%%%%%%%%%%%%%%%%%

\section{Expert Evaluation Guidelines}
\label{app:annotation}

\subsection{Task Description}

The two annotators are patent practitioners rather than academic observers of patents: one is a patent attorney with 8 years of professional experience, and the other holds a PhD in patent law. Annotators are presented with a patent description (input) and a generated claim set (output). They independently rate each output on two dimensions using a 5-point Likert scale.

\subsection{Scoring Rubric}

\paragraph{Structural Quality (1--5).}

\begin{itemize}
  \item[\textbf{5}] All claims form a valid dependency tree; every dependent claim correctly references its parent; no missing antecedent bases; proper use of transitional phrases and claim formatting throughout.
  \item[\textbf{4}] Dependency structure is mostly correct with at most one minor error (e.g., a single missing back-reference or formatting inconsistency); antecedent bases are nearly all resolved.
  \item[\textbf{3}] Overall structure is recognizable but contains multiple errors: some broken dependency chains, occasional missing antecedent bases, or inconsistent formatting across claims.
  \item[\textbf{2}] Significant structural problems: many claims lack proper back-references, dependency chains are frequently broken, or the output mixes independent and dependent claim formats incorrectly.
  \item[\textbf{1}] No discernible claim structure: output is a flat text block, lacks claim numbering, or is entirely incoherent as a patent claim set.
\end{itemize}

\paragraph{Semantic Quality (1--5).}

\begin{itemize}
  \item[\textbf{5}] Claims fully and accurately capture the inventive concept from the description; independent claims define appropriate scope; dependent claims add meaningful narrowing limitations.
  
  \item[\textbf{4}] Claims capture the core invention with minor omissions or slight scope inaccuracies; dependent claims generally narrow scope appropriately.
  \item[\textbf{3}] Claims partially capture the invention but miss key technical elements or include some limitations that contradict or are irrelevant to the description.
  \item[\textbf{2}] Claims show only superficial connection to the description; significant technical content is missing, hallucinated, or misrepresented.
  \item[\textbf{1}] Claims are unrelated to the input description, entirely hallucinated, or too incoherent to convey any inventive concept.
\end{itemize}

\subsection{Annotation Procedure}

Each annotator independently evaluates 200 randomly sampled test instances without access to the other annotator's ratings or knowledge of which model produced each output. Outputs from all four models are shuffled and presented in randomized order to prevent positional bias. Annotators are instructed to read the full patent description before rating each corresponding claim set.